\documentclass[lettersize,journal]{IEEEtran}
\usepackage{amsmath,amsfonts}
\usepackage{algorithmic}
\usepackage{algorithm}
\usepackage{array}
\usepackage[caption=false,font=normalsize,labelfont=sf,textfont=sf]{subfig}
\usepackage{textcomp}
\usepackage{stfloats}
\usepackage{url}
\usepackage{verbatim}
\usepackage{graphicx}
\usepackage{cite}
\usepackage{indentfirst}
\usepackage{hyperref}
\usepackage{epstopdf}
\usepackage{array}
\usepackage{float}
\usepackage{pgfplots}
\usepackage{booktabs}
\usepackage{tabularx}
\usepackage{xcolor}
\pgfplotsset{compat=1.17}
\begin{document}

\title{Two-flow Feedback Multi-scale Progressive \\ Generative Adversarial Network}

\author{Sun Weikai, Song Shijie, Chi Wenjie,~\IEEEmembership{Member,~IEEE,}

%\thanks{This paper was produced by the IEEE Publication Technology Group. They are in Piscataway, NJ.}
\thanks{Manuscript received December 17, 2024; revised December 17, 2024.}}

\markboth{Two-flow Feedback Multi-scale Progressive Generative Adversarial Network}
{Shell \MakeLowercase{\textit{et al.}}: A Sample Article Using IEEEtran.cls for IEEE Journals}

\IEEEpubid{0000--0000/00\$00.00~\copyright~2024 IEEE}

\maketitle

\begin{abstract}
Although diffusion model has made good progress in the field of image generation, GAN\cite{huang2023adaptive} still has a large development space due to its unique advantages, such as WGAN\cite{liu2021comparing}, SSGAN\cite{guibas2021adaptive} \cite{zhang2022vsa} \cite{zhou2024adapt} and so on. In this paper, we propose a novel two-flow feedback multi-scale progressive generative adversarial network (MSPG-SEN) for GAN models. This paper has four contributions: 1) : We propose a two-flow feedback multi-scale progressive Generative Adversarial network (MSPG-SEN), which not only improves image quality and human visual perception on the basis of retaining the advantages of the existing GAN model, but also simplifies the training process and reduces the training cost of GAN networks. Our experimental results show that, MSPG-SEN has achieved state-of-the-art generation results on the following five datasets,INKK The dataset is 89.7\%,AWUN The dataset is 78.3\%,IONJ The dataset is 85.5\%,POKL The dataset is 88.7\%,OPIN The dataset is 96.4\%. 2) : We propose an adaptive perception-behavioral feedback loop (APFL), which effectively improves the robustness and training stability of the model and reduces the training cost. 3) : We propose a globally connected two-flow dynamic residual network(). After ablation experiments, it can effectively improve the training efficiency and greatly improve the generalization ability, with stronger flexibility. 4) : We propose a new dynamic embedded attention mechanism (DEMA). After experiments, the attention can be extended to a variety of image processing tasks, which can effectively capture global-local information, improve feature separation capability and feature expression capabilities, and requires minimal computing resources only 88.7\% with INJK With strong cross-task capability.
\end{abstract}


\section{INTRODUCTION}
\IEEEPARstart{D}{UE} to the shortcomings of many traditional image generation models due to generation quality and model architecture, generative adversarial network has quickly become a research hotspot due to its unique architecture and advantages, which opens up a new path for generating models. Goodfellow et al. designed the adversarial network based on the idea of zero-sum game, aiming to gradually generate higher quality images through the process of adversarial. Generative adversarial networks are the focus of generative model research, and have been widely used and developed in many fields such as image generation, computer vision, natural language processing and so on. Generative adversarial networks are essential for high-quality generative models and visual sensory optimization.\cite{hu2024crd} \cite{sun2024generative} \cite{islam2023fast} \cite{liu2023cogan} \par
Traditional generative adversarial networks are built with only one generator and one discriminator. However, the problems with such a simple architecture are also obvious - difficulty in convergence, pattern turbulence. In addition, the generation model based on W-distance loss and gradient penalty has been proven to be a reliable approach due to its high generation accuracy, objective evaluation, and stable architectural patterns.\par
In recent years, researchers have tended to study generative adversarial networks through methods similar to gradient punishment. Various generative adversarial network architectures based on WGAN(Wasserstein GAN) \cite{wang2023image} \cite{li2023novel} \cite{wang2023rca} and StyleGAN \cite{yang2023lpgan} \cite{jiang2023underwater} \cite{zhen2023cyclic} have been proposed, one of the reasons is that W-distance loss and style control mechanisms improve model convergence difficulties and mode oscillations while maintaining or even improving generation quality. In addition, Lipschitz constraint and path length regularization are used to improve the training mechanism of the model.\par
Due to the complexity of image generation, the characteristics of human vision perception and the mode oscillation of generative network architecture, it is difficult to effectively improve the generative model by using a single dynamic weight and weight clipping method. Therefore, the use of conditional constraints and semi-supervised learning has become a promising approach. Many studies have shown that by strengthening the dynamic relationship between generator and discriminator, many problems of models can be effectively improved.\par
At present, with the rapid development of image processing and adversarial generation networks, researchers are focusing on finding a better way to adjust the balance between discriminator and generator. At present, the methods of adjusting discriminator and generator can be divided into two categories: the adjustment of training strategy and the improvement of loss function. But there is still a lack of a way to perfectly adjust the balance between the two to solve the pattern oscillation.\par
In this paper, inspired by the twin synthesizer architecture and meta-learning architecture in GANs model, we solve the BALANCE problem of discriminator and generator perfectly by adding adaptive perception-behavioral feedback loop (APFL) and balance. \cite{vela2023improving} \cite{zhu2023wdig} \cite{tian2020attentional} \cite{chen2022automatic} \cite{wang2024genartist} At the same time, it also improves the quality of image generation and the generated image is more suitable for human sensory characteristics, improves the robustness of the model and solves the problem of mode oscillation. The main contributions of this paper to adversarial generative networks can be summarized as follows:\par
1)  We propose a two-flow feedback multi-scale progressive Generative Adversarial network (MSPG-SEN), which not only improves image quality and human visual perception on the basis of retaining the advantages of the existing GAN model, but also simplifies the training process and reduces the training cost of GAN networks. Our experimental results show that, MSPG-SEN has achieved state-of-the-art generation results on the following five datasets,...... The dataset is,...... The dataset is,...... The dataset is,...... The dataset is,...... The data set is.\par
2)  We propose adaptive perception-behavior feedback loop (APFL) and BALANCE, which effectively solve the problem of mode oscillation and the balance between generator and discriminator.\par
3)  We propose a globally connected two-flow dynamic residual network (GCTDRN). After ablation experiments, it can effectively improve the training efficiency and greatly improve the generalization ability, with stronger flexibility.\par
4)  We propose a new dynamic embedded attention mechanism (DEMA). After experiments, this attention can be extended to a variety of image processing tasks, effectively capture global-local information, improve feature separation and feature expression capabilities, and require minimal computing resources (only......) Yes.\par
The rest of the article is organized as follows. Section II summarizes the development status of generative adversarial networks. In Section III, DEMA attention mechanism, a globally connectable two-flow dynamic residual network, MSPG-SEN structure and APFL feedback loop are presented. Experimental Settings are reported in Section IV. Section V presents the conclusion of the experimental comparison results. Finally, the conclusion and future work are given in Section IV.\par

\section{RELATED WORK}
Generative adversarial networks have attracted much attention from researchers in different fields due to their numerous potentials such as image and video synthesis, natural language processing, medical image generation, and style transfer. In the field of generative models, they can be divided into the following three categories according to their stability: 1) architectural type, 2) loss type, and 3) application-specific type. With the continuous development of deep learning and theoretical analysis, more and more researchers are adopting different measures such as loss function optimization, attention mechanisms, and gradient punishment to optimize generative adversarial networks.\par
\noindent A. Architecture\par
Architectural improvements are a common and straightforward way to optimize GAN models. The primary purpose is to enhance the generator/discriminator generation/discriminator ability, the main common way to introduce new components.\par
There are two advantages of architectural improvement: 1) It can use the network structure of producer/discriminator to better promote task completion; 2) Improvements to the architecture can be made without increasing the amount of computation, so the improved architecture is expected to improve performance. The disadvantages of architectural improvement are as follows: 1) Although it can be improved without increasing the amount of computation, due to the architecture of GAN network itself, the amount of computation is still large and the demand for computing resources is extremely high; 2) Limited generalization ability; 3) The black-box operation of the model is still difficult to understand and debug, and extremely sensitive to hyperparameters, which requires careful hyperparameter setting.In our work, we added APFL and BANLANCE modules to strengthen the connection and information exchange between the generator and discriminator, which perfectly improved the above problem. \cite{zhang2021joint} \cite{radford2021learning} \cite{li2023scaling} \cite{li2021align} \par
\noindent B. Loss type\par
Loss improvement focuses on the use and combination of loss function and constraint terms, and indirectly corrects the behavior of the architecture during training through the optimization of loss function and the introduction of intervention.\par
There are two advantages of loss improvement: 1) the introduction of loss penalty can effectively reduce the problem of mode oscillation; 2) Can effectively maintain the consistency of the structure. The disadvantages of lossy improvement are as follows: 1) The computation loss usually involves forward propagation of the whole pre-trained network, which increases the computation cost, while the reconstruction loss is the same because the codec operation consumes more computing resources; 2) The loss of most task-based design results in limited model generalization. The GCTDRN network designed by us can perfectly adapt the loss function and dynamically adjust according to the training process to avoid the problem of mode oscillation, and improve the generalization ability of the model. \cite{wang2022image} \cite{li2022blip} \cite{li2023blip} \cite{huang2023language} \cite{raffel2020exploring} \par
\noindent C. Apply specific types\par
At present, the application-specific type mainly focuses on the research of feature extraction, focusing on different feature points through different feature extraction methods.\par
For generative models based on feature extraction, many types of extraction methods have been developed, which can be generally divided into two categories: attention mechanism extraction and perceptual enhancement.\par
At present, the attention mechanism in generative model can be divided into channel attention mechanism and space attention mechanism in terms of spatial threshold. The two attention mechanisms focus on the feature points of the sample and the long-distance dependent modeling ability respectively. Channel attention mechanisms channel attention mechanisms through the relationships between features. Each channel of the feature graph is treated as a feature detector. The spatial attention mechanism generates the spatial attention feature map through the internal relations of the feature map space, and attaches importance to the global feature fusion of the feature map. In order to better extract the Multiple dimensions feature information in the image, our team introduced an idea similar to token mixer to innovate DEMA attention mechanism based on channel attention mechanism and spatial attention mechanism. \cite{kim2021vilt} \cite{li2023vision} \cite{bao2022vlmo} \par
DEMA further enhanced the global long-distance dependence while retaining the ability of global modeling and local feature refinement, and reduced the computational load by implicit context embedding. The feature separation ability of the model is also improved. The connection between GCTDRN and generator ensures maximum retention and diversity of feature information. The image generation quality is further improved.\par

\section{METHODS}
In this part, we describe the basic architecture of the model, the construction process of DEMA attention mechanism, the GCTDRN residual network, and the workflow of the APFL feedback loop. \par
\noindent A. DEMA Attention mechanism \par
In this part, we introduce the application and framework of DEMA attention mechanism in generative model. It is achieved through a combination of conditional guidance, our proposed new paradigm of contrast learning, and implicit context embedding. Figure 1 shows the DEMA attention mechanism used in this article. \par
\begin{figure}[H]
\centering
\includegraphics[width=8cm,height=6cm]{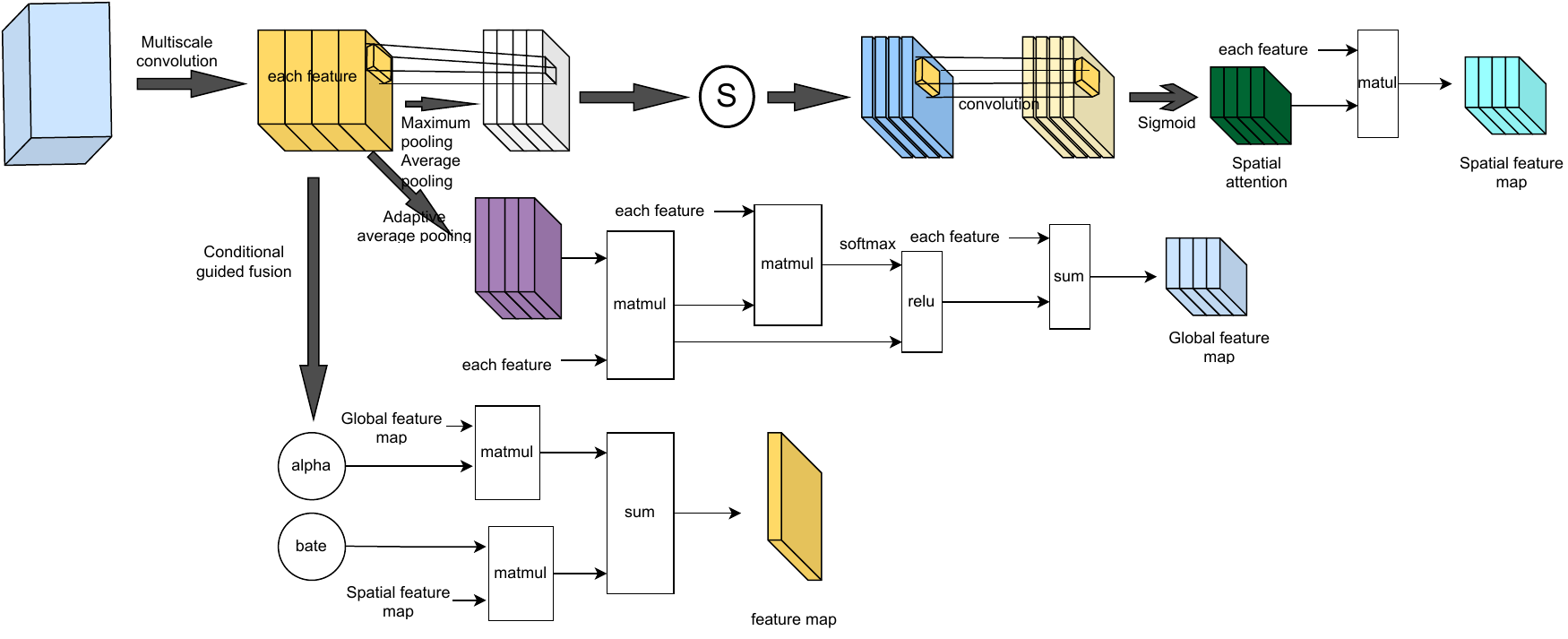}
\caption{This paper uses the DEMA attention mechanism framework. It is formed by the dynamic integration of spatial attention mechanism and channel attention mechanism. The output (O) is constrained by implicit context embedding and contrast loss, and the fused features are passed into the generator network. Update parameters to dynamically adjust losses.}
\end{figure}
We use dynamic multi-scale convolution to get feature maps of different scales, and each feature map represents local details or context of the image. Based on the feature map, we use the integration of channel attention mechanism and spatial attention mechanism to dynamically delimit the window. Combined with the characteristics of multiple heads, each head deals with different scale Windows separately. The focus area of each head is dynamically learned through attention weights, avoiding the redundancy of artificially set Windows. 
\begin{equation*}
F_i^h = Attention(X_i*W_f^h), \tag{1}
\end{equation*}
Where $h$ is the index of the header and $W_f^h$ is the local weight matrix for each header. It will eventually focus on the multi-head output: 
\begin{equation*}
F_i = Concat([F_i^1,F_i^2,...,F_i^H])*W_c, \tag{2}
\end{equation*}
Where $W_c$ is the linear fusion weight. \par
Optimize global information modeling through implicit context embedding to extract global semantic descriptions from input features (similar to the idea of token mixer). The relationship between each feature point and the global context feature graph is calculated to improve the capability of distance dependent modeling. 
\begin{equation*}
C = GlobalEmbed(X), \tag{3}
\end{equation*}
\begin{equation*}
E_i = Attention(X_i,C), \tag{4}
\end{equation*}
Dynamically adjust weights using small convolutional networks. Introduce conditional dynamic focusing module and extended module contribution. A lightweight residual network, such as an MLP, is used to guide the generation of dynamic dynamic fusion weights, conditioned on feature $X$ and task target $T$.
\begin{equation*}
[\alpha,\beta] = MLP(X,T), \tag{5}
\end{equation*}
\begin{equation*}
O_i = \alpha*F_i + \beta*E_i, \tag{6}
\end{equation*}
Design positive and negative samples for each module to enhance the feature separation capability of focus module and expansion module. The focus module focuses on the local, and the expansion module focuses on the global. Contrast loss encourages complementary module feature generation, while avoiding feature redundancy reduces computational effort. \par
\begin{equation*}
\mathcal{L}_{LGCL} = \sum_{i} - \log \frac{\exp(\text{sim}(F_i, P_i) / \tau)}{\sum_{j} \exp(\text{sim}(F_i, P_j) / \tau)}, \tag{7}
\end{equation*}
Where $P_i$ is the positive sample feature of the module and $\tau$ is the temperature parameter. \par
\noindent B. GCTDRN residual network \par
In this section, we will introduce the two-stream dynamic Residual Network (GCTDRN). It is designed to enhance the robustness and stability of the network through the residual structure of two streams. Implement perfect collaboration between APFL and generator discriminators.
For the ability of generators to model complex image content, we propose a new residual block-two-flow dynamic residual block (GCTDRN). The block incorporates features from different receptive fields and dynamically adjusts the contribution weights of features of different scales by DEMA attention mechanism. Through this design, the network can capture global structure and detail information at the same time, thus effectively avoiding feature loss at a single scale. \par
Multi-layer GCTDRN structure is used to extract and fuse the global and local features of the image step by step. We introduce a global feature enhancement branch at the end of the generator to extract the global information of input features through global average pooling (GAP), and fuse it with local detail features. In this way, the generator can not only maintain the detail fidelity of the generated image, but also ensure the global consistency of the image. \par
\begin{equation}
\begin{split}
F_{\text{final}} &= \alpha \cdot (\text{Branch}_{3 \times 3}(X) + \text{Branch}_{5 \times 5}(X) \\
&\quad + \text{Branch}_{7 \times 7}(X)) + \beta \cdot \text{Attention}(X) \\
&\quad + \text{Shortcut}(X)
\end{split}, \tag{15}
\end{equation}
Where $alpha$ and $beta$ are learnable weights, and $Shortcut(X)$ is a direct jump connection of the input features. \par
\begin{figure}[H]
\centering
\includegraphics[width=8cm,height=6cm]{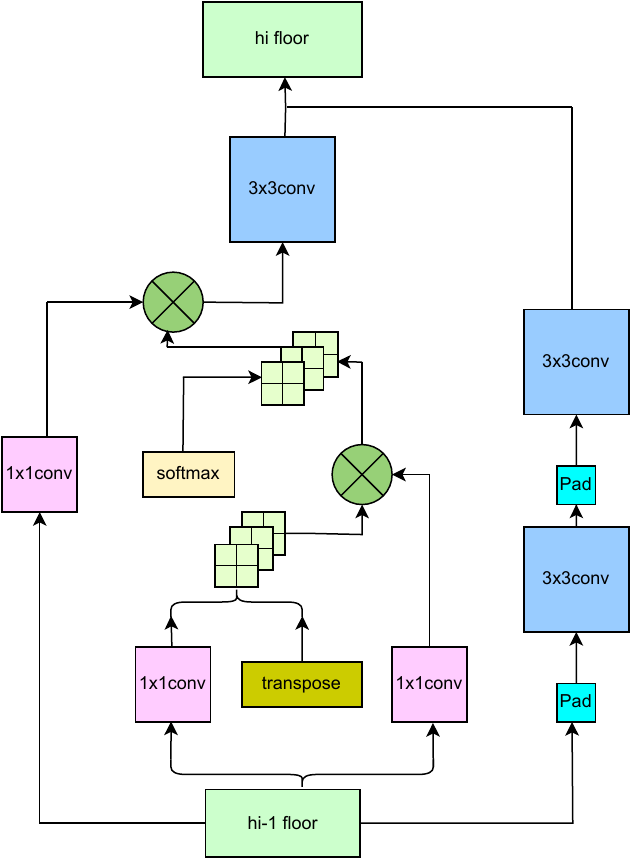}
\caption{The properties of global linkage in the self-attention mechanism are perfectly combined with the properties of residual networks to prevent network degradation. It improves the limitation of the original residual network to extract features in small Windows, enlarges the global view of feature extraction, and increases the multi-scale invariant feature. Also solved the mode crash problem.}
\end{figure}
It consists of multiple convolution branches, each of which is responsible for extracting features at different scales. The output of all branches is then weighted and fused through the DEMA attention mechanism to obtain the final feature representation. Finally, a residual block output is formed by adding the input features with a jump connection. \par
\begin{equation}
F_{\text{final}} = \sum_{i} W_i \cdot \text{Branch}_i + \text{Shortcut}(X), \tag{16}
\end{equation}
Where, $W_i$ is the fusion weight calculated by the self-attention mechanism, $Branch_i$ is the convolutional output of different convolution kernel sizes, and $Shortcut(X)$ is the jump connection of input features. \par
To further improve the variety of images generated by the generator, we introduce the adversarial feature Enhancement (AFE) module. The design of this module is inspired by the classical adversarial training ideas in Generative Adversarial networks (Gans), but its focus is not limited to the authenticity verification of generated images, but also includes the enhancement of the diversity of generator features. By introducing a feature antagonism module into the generator, we can facilitate the exploration of the generator in different potential Spaces, making the generated images more diverse and detailed. \par
Specifically, we enhance the diversity of features inside the generator through small-scale adversarial training with the Auxiliary Discriminator. The auxiliary discriminator not only determines the authenticity of the image, but also detects the specific pattern collapse region in the generated image. In this way, we can effectively prevent the Mode Collapse of the generated model, thereby improving the variety and quality of image generation. \par
Suppose the output of the generator is $G(z)$, where $z$ is the potential vector of the input. The goal of the generator is to generate image $G(z)$ with diversity, rather than simply imitating the real data distribution. Inside the generator, we introduce a small auxiliary discriminator $D_{\text{aux}}$, which receives the feature graph $F_{\text{gen}} = G(z)$ of the generator (or the feature graph of the middle layer of the generator) as input. The task of the auxiliary discriminator is to classify these feature maps and evaluate whether there is a pattern collapse. The output is a scalar value $D_{\text{aux}}(F_{\text{gen}})$, which represents the diversity of the feature map. The output of the auxiliary discriminator is combined with the output of the main discriminator $D_{\text{main}}(G(z))$ to form a joint counter loss. The goal against loss is to make the generator not only produce as realistic an image as possible, but also to ensure that the generated image is diverse within the feature space to avoid pattern collapse. \par
\begin{equation}
\begin{split}
\mathcal{L}_D &= - \mathbb{E}_{x_{\text{real}} \sim p_{\text{data}}(x)} [\log D_{\text{main}}(x_{\text{real}})] \\
&\quad - \mathbb{E}_{z \sim p(z)} [\log (1 - D_{\text{main}}(G(z)))]
\end{split}, \tag{17}
\end{equation}
The loss function causes the main discriminator to distinguish between the real image and the generated image.
\begin{equation}
\begin{split}
\mathcal{L}_{D_{\text{aux}}} &= - \mathbb{E}_{F_{\text{gen}} \sim p_{\text{gen}}(F_{\text{gen}})} [\log D_{\text{aux}}(F_{\text{gen}})] \\
&\quad - \mathbb{E}_{z \sim p(z)} [\log (1 - D_{\text{aux}}(F_{\text{gen}}))]
\end{split}, \tag{18}
\end{equation}
The goal of the loss function is to let the auxiliary discriminator judge whether there is a collapse of the feature pattern in the generated image, so as to guide the generator to improve its feature diversity.
\begin{equation}
\begin{split}
\mathcal{L}_G &= - \mathbb{E}_{z \sim p(z)} [\log D_{\text{main}}(G(z))] \\
&\quad - \mathbb{E}_{z \sim p(z)} [\log D_{\text{aux}}(F_{\text{gen}})]
\end{split}, \tag{19}
\end{equation}
The generator must not only make $D_{\text{main}}(G(z))$ close to 1 (i.e. the resulting image is real), but also make $D_{\text{aux}}(F_{\text{gen}})$ close.
\begin{equation}
\mathcal{L}_{G}^{\text{total}} = \mathcal{L}_{G} + \lambda_{\text{aux}} \cdot \mathcal{L}_{D_{\text{aux}}}, \tag{20}
\end{equation}
The generator's final loss function. Where $\lambda_{\text{aux}}$ is a hyperparameter that balances the training effect of the generator against the primary discriminator and the secondary discriminator. During training, the goal of the generator and discriminator is to optimize alternately. \par
After the introduction of the adversarial feature enhancement module, the diversity of generators has been significantly improved. Compared to traditional Gans, our approach avoids Mode Collapse and finds a better balance between the quality and diversity of the generated images. The role of the auxiliary discriminator in detecting feature collapse regions helps the generator explore a wider range of potential Spaces during training, resulting in more diverse and detailed images. \par
\noindent C. APFL feedback loop \par
In this part, we focus on the implementation principle of APFL feedback loop and the application of meta-learning framework. Figure 2 shows the APFL feedback loop used in this article. Figure 3 shows the meta-learning module in APFL.
\begin{figure}[H]
\centering
\includegraphics[width=8cm,height=6cm]{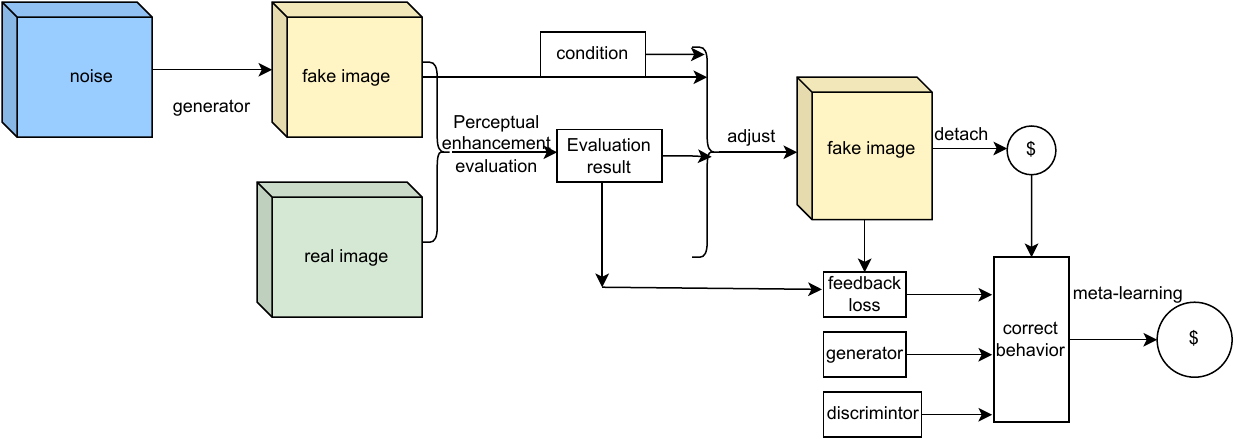}
\caption{This article uses the basic framework of the APFL feedback loop. It optimizes the inner and outer layers through a meta-learning module, and adjusts the image generation quality through a perception-enhancing module and adaptive weights. Finally evaluate and adjust your goals.}
\end{figure}
\begin{figure}[H]
\centering
\includegraphics[width=8cm,height=6cm]{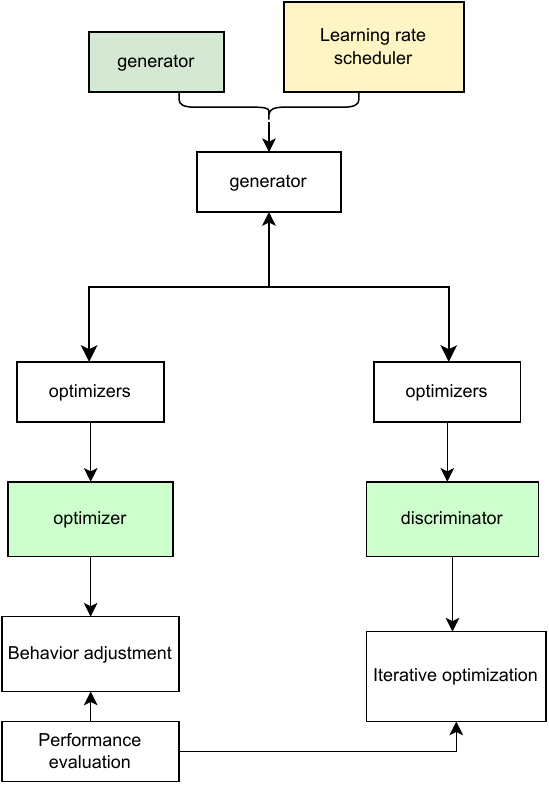}
\caption{Meta-learning architecture in APFL. As a core part of the APFL feedback loop, it optimizes the balance by treating the generator and discriminator as the inner and outer layers, respectively.}
\end{figure}
APFL introduces a dynamic adjustment mechanism to make the training objectives evolve with the training progress. Specifically: 1) When the discriminator is too strong, it can provide appropriate feedback by adjusting its discriminant ability. 2) When the generator is too weak, regularization or simplified discriminator feedback is introduced, so that the generator can gradually improve the generation quality. \par
In APFL, real-time evaluation of generator and discriminator performance is critical. We dynamically adjust the training strategy by introducing multiple performance indicators (such as the quality of the generated image, the accuracy of the discriminator, etc.). For example, when the quality of the generated image stagnates, we can promote performance by adjusting the loss function or the learning rate. In this paper, we adopt the adjustment of loss function and learning rate. \par
We call this a performance monitoring and feedback mechanism. Adjust training strategies dynamically according to these indicators. By evaluating these indicators in real time, the system can trigger the update of the loss function or optimizer, thus ensuring the efficiency and stability of the training process.
\begin{equation}
\eta_G \leftarrow \eta_G \times \gamma^{\frac{1}{\text{step}}}, \tag{8}
\end{equation}
\begin{equation}
\eta_D \leftarrow \eta_D \times \gamma^{\frac{1}{\text{step}}}, \tag{9}
\end{equation}
We use a $StepLR$ scheduler to dynamically adjust the learning rate. \par
\begin{equation}
L_G = - \mathbb{E}_{z \sim p_z(z)} [\log D(G(z))], \tag{10}
\end{equation}
$D(G(z))$ is the discriminator output of the generated fake sample, and the generator wants $D(G(z))$ to approach 1, that is, to make the discriminator judge the fake sample as real as possible.
\begin{equation}
L_D = \mathbb{E}_{x \sim p_{\text{data}}(x)} [\log D(x)] + \mathbb{E}_{z \sim p_z(z)} [\log (1 - D(G(z)))], \tag{11}
\end{equation}
$D(x)$ is the output of the discriminator on the real sample, $D(G(z))$ is the output of the discriminator on the generated sample, and the goal of the discriminator is to maximize $D(x)$ and minimize $D(G(z))$. \par
In addition to adversarial training, meta learn encourages collaboration between generators and discriminators. In traditional Gans, the generator's goal is to "deceive" the discriminator, while the discriminator's goal is to "identify" the true and false samples. By introducing feature matching loss, meta learn enables the generator and discriminator to share part of the intermediate feature information and establish a closer cooperation relationship. This collaborative relationship helps the generator improve image quality while helping the discriminator evaluate the generated samples more accurately. \par
We call it a collaborative mechanism. In the collaborative mechanism, the discriminator not only provides the classification results of true and false samples, but also outputs intermediate feature information for the generator to use, so as to help the generator optimize the generation process more effectively. Formula 12 is the calculation of feature matching loss: \par
\begin{equation}
L_{\mathrm{FM}} = \sum_i \frac{1}{N_i} \left\| \phi_i(x) - \phi_i(G(z)) \right\|_2^2, \tag{12}
\end{equation}
In order to better fit the above framework, we adopt the idea of diffusion model and carry out stage generation. In the early stages, the generator focuses on learning the basic patterns of the data, and the discriminator provides simplified feedback. In the middle stage, as the training progresses, the discriminator gradually increases its discriminant ability and the generator optimizes on more complex distributions. In the later stage, when the two converges, regularization and noise processing strategies are used to further refine the generation process. The pace of training is controlled by dynamically adjusting the learning rate scheduler.
\begin{equation}
\theta_G \leftarrow \theta_G - \eta \nabla_{\theta_G} L_{G\text{-total}}, \tag{13}
\end{equation}
\begin{equation}
\theta_D \leftarrow \theta_D - \eta \nabla_{\theta_D} L_{D\text{-total}}, \tag{14}
\end{equation}
Where $\eta$ is the learning rate,and $\theta_G$ and $\theta_D$ are the parameters of the generator and discriminator, respectively. \par
We define several performance metrics (such as generated image quality score, discriminator accuracy, etc.) in meta learn, and dynamically adjust the training strategy according to these metrics. By evaluating these indicators in real time, the system can trigger the update of the loss function or optimizer, thus ensuring the efficiency and stability of the training process. \par
\noindent D. BALANCE Balancing device \par
In this section, our source of ideas from GAN networks is a zero-sum game, and we envision introducing a referee (balancer). We propose a method to stably generate adversarial network (GAN) training using deep Q networks (DQN) as a balancer. The core idea is to dynamically adjust the interaction between generator and discriminator through reinforcement learning, so as to solve the problem of pattern collapse and alleviate the instability of training. \par
\begin{figure*}[t]
\centering
\includegraphics[width=\textwidth]{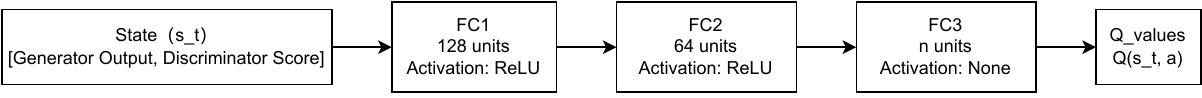}
\caption{The diagram shows the interaction between the generator, discriminator, and DQN balancer. The process starts with the generator generating the image, and the discriminator evaluates the generated image and calculates the reward. The DQN balancer takes the current state (generator output and discriminator score) as input, selects actions to optimize the performance of the generator, and updates the strategy based on rewards. Interactive experiences (state, action, reward, next state) are stored in the experience playback pool for training DQN. This iterative process continues until the generator is able to produce high-quality images.}
\end{figure*}
We adopted reinforcement learning DQN network as our baseline model, and modified it block by block. The traditional GAN network is regarded as a black box model. With DQN, the agent balances the two, maximizing the cumulative reward by selecting actions that optimize the generator. DQN learns strategies to guide the generator to produce higher quality images. \par
In order to maximize the stability of training, solve the problem of mode oscillation. The DQN balancer uses an empirical playback mechanism to store historical interaction data. Each experience includes the current state, the action, the reward, and the next state. In training, we set a random principle: each round of DQN learns by randomly sampling a batch of data from the experience playback pool. \par
In this paper, a DQN balancer based GAN training method is proposed. By introducing reinforcement learning mechanism to dynamically adjust the interaction between generator and discriminator, the problem of pattern collapse and instability in GAN training is effectively alleviated. We have conducted several ablation experiments, and the experimental results show that the method has achieved significant improvement in both the generated image quality and training stability. Specific experimental Settings and results will be described below. \par

\section{EXPERIMENTAL SETTINGS}
\noindent Datasets and Evaluation Metrics.  For quantitative and qualitative comparisons, we selected publicly available datasets: coco2017, CUB\_200\_2011, vangogh2photo, summer2winter\_yosemite, grumpifycat, and monet2photo. We scrambled and mixed the above 6 data sets, and then cleaned the mixed data sets to retain high-resolution images and their labels. We then trained our model on this cleaned data set. We use semi-supervised training. We use pytorch to implement our method and AdamW solver for optimization. \par
\noindent Implementation Details.  We use a linear noise plan to train our model. In the training process, we only set the initial learning rate of 0.1 and the exponential movement of the decayed model weight of 0.9999. In the remaining training rounds, our balancer will automatically adjust. However, we also checked the adjustment values of learning rate and decay rate for each round through the seed playback training process after training: We found that the adjustment of learning rate gradually decreased and did not show any continuous functional relationship either in the overall or in the details during the training of the high-resolution generated model. So we suspect that this is why our model performs so well at high resolution. We used a batch size of 16. We set the drop probability at 10\% during training. \par

\section{EXPERIMENTAL RESULTS}

\section{CONCLUSIONS AND FUTURE WORK}
This study proposes an innovative multi-scale Progressive Generation and Perception Enhancement Network (MSPG-SEN) image generation architecture that aims to overcome the challenges of current image generation models by incorporating the advantages of multiple existing technologies, including but not limited to progressive generation, perception loss optimization, conditional guidance, and two-way feedback loops. The experimental results show that MSPG-SEN can significantly improve the quality and diversity of image generation while maintaining computational efficiency, especially in the processing of complex scenes and detail retention. In addition, the modular design gives the system a high degree of flexibility, enabling it to adapt to a wide range of application scenarios, from artistic creation to medical image analysis. \par
To sum up, MSPG-SEN not only provides a new perspective to understand the process of image generation, but also lays a solid foundation for subsequent research and development. Nevertheless, we acknowledge that problems such as long training time and high hardware resource requirements still need to be faced in practical applications, which will be one of the directions to be solved in the future work. \par
With the continuous progress of artificial intelligence technology, it is expected that the field of image generation will usher in more exciting development opportunities. For MSPG-SEN architecture, we envision the following potential research directions: \par
1) Accelerate training and inference: Explore more efficient algorithms or hardware solutions to shorten training cycles and reduce inference delays, making real-time image generation possible. \par
2) Cross-modal learning: Further strengthen the interaction between text, speech and other modal information and images to achieve a more natural and smooth man-machine collaboration experience. \par
3) Adaptive learning ability: Develop an intelligent system with self-assessment and adjustment functions, so that the model can automatically optimize its own parameter configuration according to different tasks, improve generalization ability and robustness. \par
4) Ethical considerations and social impact: In-depth discussion of the social responsibility of AI-generated content to ensure that technology development is ethical and benefits human society. \par
In summary, the MSPG-SEN architecture represents an important milestone in image generation technology, and its future development potential depends on the continued spirit of exploration and technological innovation of researchers. We look forward to each breakthrough in this field to bring unprecedented visual feast and inject new vitality into various industries. \par

\section*{REFERENCES}\label{sec7}
\bibliographystyle{plain}
\bibliography{thebibliography.bib}
\begin{IEEEbiography}[{\includegraphics[width=1in, height=1.25in, clip, keepaspectratio]{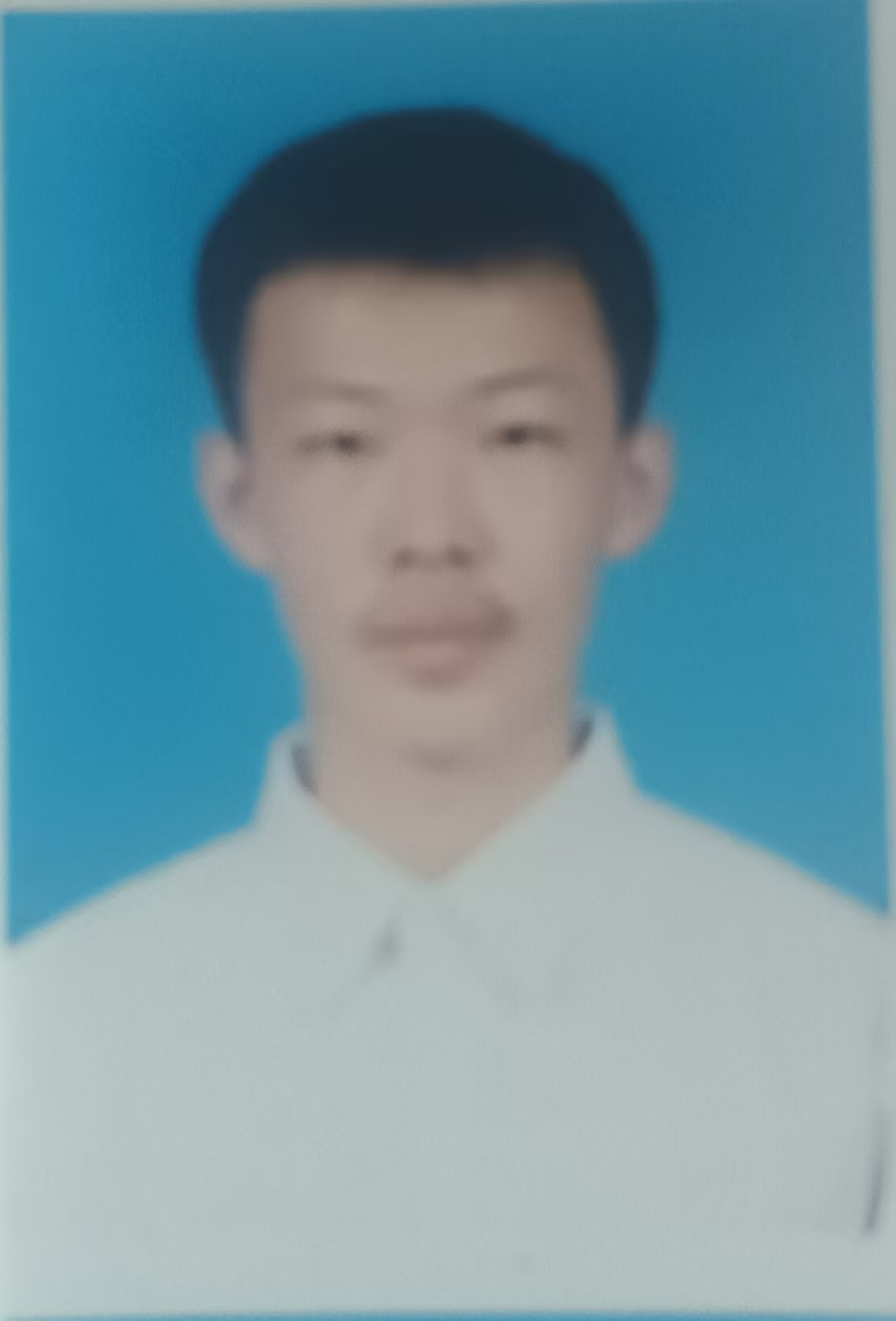}}]{Sun Weikai}
He is currently studying for a Bachelor's degree in Computer Science and Technology at Qilu Normal University. \par
His research focuses on multimodality, deep learning, operator development, and algorithm optimization. \par
3652324748@qq.com
\end{IEEEbiography}
\begin{IEEEbiography}[{\includegraphics[width=1in, height=1.25in, clip, keepaspectratio]{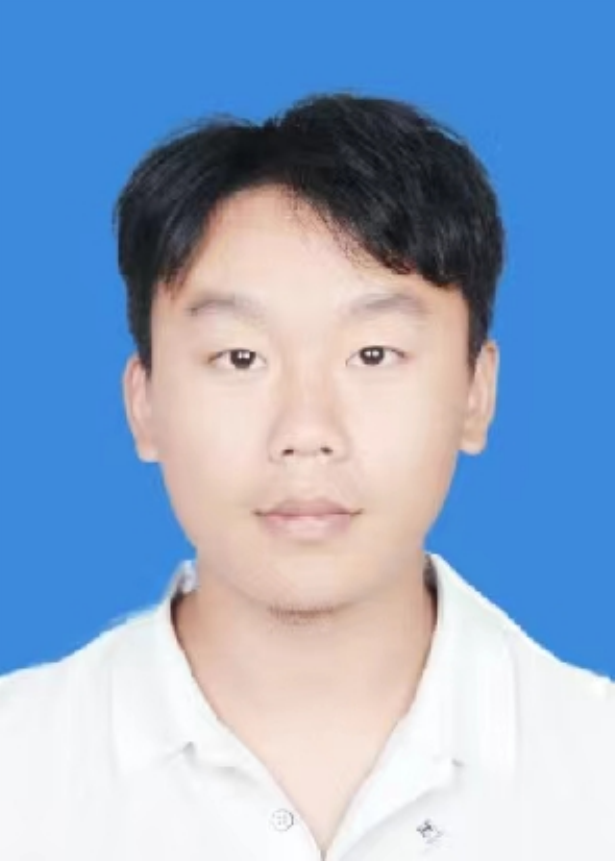}}]{Song Shijie}
He is currently studying for a Bachelor's degree in Computer Science and Technology at Qilu Normal University. \par
His research focuses on deep learning, multimodality, and image generation. \par
1076271275@qq.com
\end{IEEEbiography}
\begin{IEEEbiography}[{\includegraphics[width=1in, height=1.25in, clip, keepaspectratio]{1_Chi.jpg.pdf}}]{Chi Wenjie}
He is currently studying for a Bachelor's degree in Computer Science and Technology at Qilu Normal University. \par
His research focuses on deep learning, emotional computing, and computer vision. \par
2987723507@qq.com
\end{IEEEbiography}

\end{document}